\documentclass[runningheads]{llncs}
\usepackage{graphicx}
\usepackage{comment}
\usepackage{amsmath,amssymb} 
\usepackage{color}

\usepackage{graphicx}
\usepackage{color}
\usepackage{longtable}
\usepackage{booktabs,subcaption,amsfonts,dcolumn}
\usepackage{bm}
\usepackage{amsmath}
\usepackage[ruled,vlined]{algorithm2e}
\usepackage{lipsum}
\usepackage{mathtools}
\usepackage{cuted}
\usepackage{flushend}
\usepackage{gensymb}
\usepackage{float}

\newcommand{\eg}{\textit{e.g.},}
\newcommand\blfootnote[1]{%
  \begingroup
  \renewcommand\thefootnote{}\footnote{#1}%
  \addtocounter{footnote}{-1}%
  \endgroup
}

\begin{document}
\pagestyle{headings}
\mainmatter


\title{Detecting Scatteredly-Distributed, Small, and Critically Important Objects in 3D Oncology Imaging via Decision Stratification} 

\titlerunning{Detecting Scatteredly-Distributed, Small, and Critically Important Objects in 3D Oncology Imaging via Decision Stratification}
%
\author{Zhuotun Zhu\inst{1,2}\and Ke Yan$^\dag$\inst{1}\and Dakai Jin$^\dag$\inst{1}\and Jinzheng Cai\inst{1}\and Tsung-Ying Ho\inst{3}\and Adam P Harrison\inst{1}\and Dazhou Guo\inst{1}\and Chun-Hung Chao\inst{4}\and Xianghua Ye\inst{5}\and Jing Xiao\inst{6}\and Alan Yuille\inst{2}\and Le Lu\inst{1}}
\authorrunning{Z. Zhu et al.}
%
\institute{
PAII Inc., Bethesda MD, USA \and Johns Hopkins University, Baltimore MD, USA
\and
Chang Gung Memorial Hospital, Linkou, Taiwan, ROC \and
National Tsing Hua University, Hsinchu City, Taiwan, ROC\and
The First Affiliated Hospital Zhejiang University, Hangzhou, China
\and 
Ping An Technology, Shenzhen, China}
\maketitle


\begin{abstract}
Finding and identifying scatteredly-distributed, small, and critically important objects in 3D oncology images is very challenging. We focus on the detection and segmentation of oncology-significant (or suspicious cancer metastasized) lymph nodes (OSLNs), which has not been studied before as a computational task. Determining and delineating the spread of OSLNs is essential in defining the corresponding resection/irradiating regions for the downstream workflows of surgical resection and radiotherapy of various cancers. For patients who are treated with radiotherapy, this task is performed by experienced radiation oncologists that involves high-level reasoning on whether LNs are metastasized, which is subject to high inter-observer variations. In this work, we propose a divide-and-conquer decision stratification approach that divides OSLNs into tumor-proximal and tumor-distal categories. This is motivated by the observation that each category has its own different underlying distributions in appearance, size and other characteristics. Two separate detection-by-segmentation networks are trained per category and fused. To further reduce false positives (FP), we present a novel global-local network (GLNet) that combines high-level lesion characteristics with features learned from localized 3D image patches. Our method is evaluated on a dataset of 141 esophageal cancer patients with PET and CT modalities (the largest to-date). Our results significantly improve the recall from $45\%$ to $67\%$ at $3$ FPs per patient as compared to previous state-of-the-art methods. The highest achieved OSLN recall of $0.828$ is clinically relevant and valuable. 

\end{abstract}

\blfootnote{$^\dag$ equal contribution.}

\section{Introduction}
Measuring lymph node (LN) size and assessing its status are important clinical tasks, usually used to monitor cancer diagnosis and treatment responses and to identify treatment areas for radiotherapy. According to the Revised RECIST guideline~\cite{eisenhauer2009new,schwartz2009evaluation}, only enlarged LNs with a short axis more than $10{\textrm{-}}15$ mm in computed tomography (CT) images should be considered as abnormal. Such enlarged LNs have been the only focus, so far, of LN segmentation and detection works~\cite{bouget2019semantic,feulner2013lymph,kitasaka2007automated,liu2016mediastinal,nogues2016automatic,oda2018dense,roth2015improving,roth2014new}. However, in cancer treatment, besides the primary tumor, all metastasis-suspicious LNs are required to be treated. This includes the enlarged LNs, as well as smaller ones that are associated with a high positron emission tomography (PET) signal or any metastasis signs in CT. This larger category is regarded as oncology significant lymph nodes (OSLNs). Identifying the OSLNs and assessing their spatial relationship and causality with the primary tumor is a key requirement for a desirable cancer treatment outcome~\cite{National2020}.

\begin{figure}[t]
\centering
\includegraphics[width=1.0\textwidth]{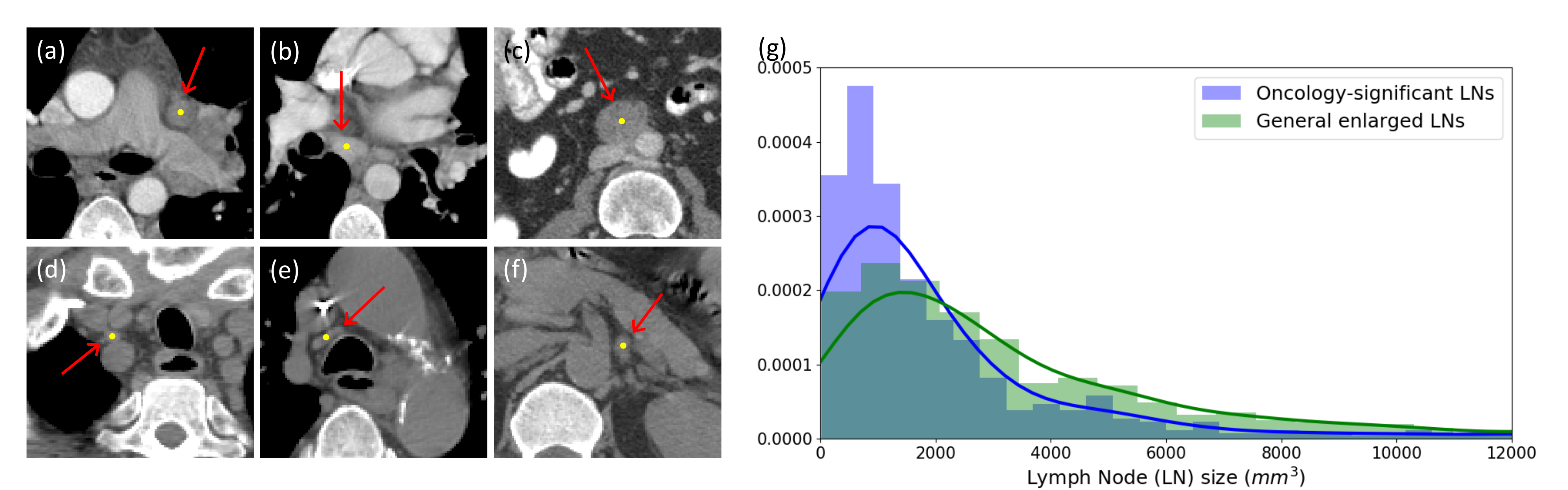}
\caption{(a,b,c) Three examples of enlarged LNs, which all prior work targets, in  \textit{contrast-enhanced} CT. (d,e,f) Three instances of OSLNs, which our work focuses on, in non-contrast RTCT. This category has not been studied before as a computational task. (g) LN volume distributions for enlarged LNs from a public dataset \cite{roth2015improving,roth2014new} and OSLNs in our radiotherapy dataset.}\label{Fig:LNSamples}
\end{figure}

Identifying OSLNs can be a daunting and time-consuming task, even for experienced radiation oncologists. It requires using high-level sophisticated reasoning protocols and faces strong uncertainty and subjectivity with high inter-observer variability~\cite{goel2017clinical}. To the best of our knowledge, this problem has not been previously tackled in a fully automatized way.  Our task on OSLNs detection is more challenging for the following reasons: (1) Finding OSLNs is often performed using radiotherapy CT (RTCT), which, unlike diagnostic CT, is not contrast-enhanced. (2) OSLNs exhibit low contrast with surrounding tissues and can be easily confused with other anatomical structures, {\it{e.g.}}, vessels and muscles, due to shape and appearance ambiguity. (3) The size and shape of OSLNs can vary considerably, and OSLNs are often scatteredly distributed at small size in a large spatial range of anatomy locations. See Fig.~\ref{Fig:LNSamples} for an illustration of the differences in appearance and size distribution between enlarged LNs the larger category of OSLNs. We can observe that OSLNs have higher frequencies at smaller sizes, challenging their detection. While, many previous works proposed automatic detection systems for enlarged LNs in contrast-enhanced CT  \cite{barbu2011automatic,bouget2019semantic,feulner2013lymph,nogues2016automatic,roth2015improving,roth2014new,yan2018deeplesion}, no work, as of yet, has focused on OSLN detection on non-contrast RTCT. Given the considerable differences between enlarged LNs and OSLNs, further innovation is required for robust and clinically useful OSLN detection. 
\begin{figure}[t]
\centering
\includegraphics[width=0.95\textwidth]{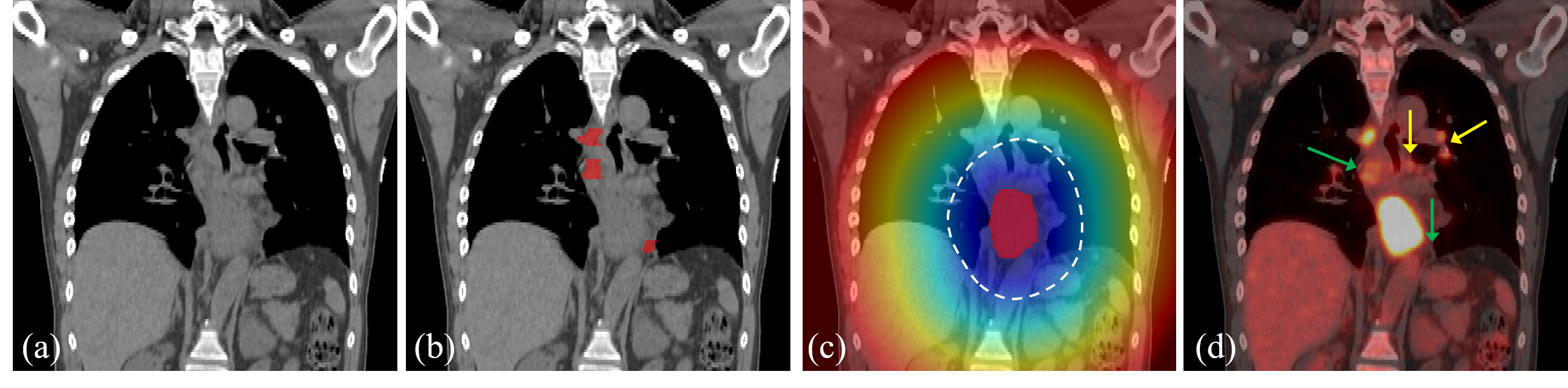}
\caption{(a) A coronal view of RTCT for an esophageal cancer patient. (b) The manual annotated OSLN mask. (c) Tumor distance transform map overlaid on RTCT. The primary tumor is indicated by red mask in the center and the white dash line shows an example of the tumor proximal and distal region division. (d) PET imaging overlaid on RTCT. The yellow arrows show several FP PET signals, and the green arrows indicate two FN OSLNs where PET has weak or even no signals. A big central bright region in PET is the primary tumor region.} \label{Fig:LNDistance}
\end{figure}

Current clinical practices offer valuable insight in how to tackle this problem. For instance, physicians condition their analysis of suspicious areas based on their distance to the primary tumor. For LNs proximal to the tumor, physicians will more readily identify them as OSLNs for the radiotherapy treatment. However, for LNs far away from the tumor, physicians are more discriminating, only including them if there are clear signs of metastasis, such as enlarged in size, increased PET signals, and/or other CT-based evidence~\cite{scatarige1983low}. Hence, distance to the primary tumor plays a key role in physician's decision making. Besides the distance, the PET modality is also highly important, as it significantly increases sensitivity~\cite{goel2017clinical}. However, PET is noisy and increased PET signals can often associate to normal physiological uptake. Moreover, PET only highlights $\sim 33\%$ of the OSLNs~\cite{leong2006prospective}. This is demonstrated in Fig.~\ref{Fig:LNDistance}(d)'s example, where PET provides key information in identifying OSLNs, which might be too difficult to detect from RTCT only. Yet, the PET also exhibits false positives (FPs) and false negatives (FNs). Based on this observation, an effective method to leverage the complementary information in RTCT and PET is crucial, but this must be done with care.

To solve this problem, we emulate and disentangle the above practices. First, we propose and validate an intuitive and effective strategy that uses distance stratification to decouple the underlying OSLN distributions into two ``tumor-proximal" and ``tumor-distal" categories, followed by training separate networks to fit the class specific imaging features to the task. LNs that are spatially close to the primary tumor site are more suspicious (even if they are not enlarged); whereas spatially distal OSLNs may need to be identified with both CT and PET imaging evidence. This type of decision uncertainty stratification is evident in medical diagnosis and our work is one of the first computational realizations. 
Second, for each OSLN category, we implement a 3D detection-by-segmentation framework that fuses predictions from two independent sub-networks, one trained on the RTCT imaging alone and the other learned via the early fusion (EF) of three channels of RTCT, PET and the 3D tumor distance map (Fig.~\ref{Fig:LNDistance}(c)).  
RTCT depicts anatomical structures, which captures  intensity appearance and contextual information, serves as a good baseline diagnostic imaging modality. In contrast, the EF stream takes into account PET's metastasis functional sensitivities as well as the tumor distance encoded in the distance transform map, which are both noisy but informative. Along with the distance stratification, this produces four predictions, which are all fused together as a late fusion (LF). This produces OSLN predictions that achieve sufficiently high sensitivities in finding OSLNs, which complements the high specificity but low sensitivity of human observers~\cite{goel2017clinical}. Missing true OSLNs can cause oncologically critical areas to remain untreated. 
Third, we propose a global-local network (GLNet) to further reduce the FP OSLN candidates obtained from above. The GLNet has two modules, with each module corresponding to the global or local spatial context. (1) For local context, we crop out any OSLN candidate region with certain context margins and adopt 3D residual convolutions~\cite{he2016deep,girshick2015fast} to extract instance-wise localized deep feature maps. 
(2) For global context, we leverage the ontology-based medical knowledge from the large-scale NIH DeepLesion~\cite{yan2018deeplesion} dataset via a lesion tagging module~\cite{yan2019holistic}, which provides high-level semantic information such as body part and shape/texture/size attributes that cannot be easily captured from local 3D image patches. The strategy of looking at locally (i.e., the imaging space) and globally (i.e., the semantic ontology space) is essential to mimic sophisticated clinical reasoning protocols. Both the imaging texture and appearance and semantically meaningful attributes are crucial to allow our workflow to filter out FPs while keeping sensitivities high.
Our contributions can be summarized as follows:
\begin{itemize}
\item To the best of our knowledge, we are the first to address the clinically critical task of detecting, identifying and characterizing OSLNs. 
\item 
We propose a novel 3D distance stratification strategy to divide and conquer the complex distribution of OSLNs into tumor-proximal and tumor-distal classes, to be solved separately, which emulates physician's decision process.
\item Besides RTCT, we incorporate the PET imaging modality and 3D tumor distance maps into a two stream detection-by-segmentation network.

\item We propose a novel GLNet to incorporate high-level ontology-derived semantic attributes of OSLNs with localized features computed from RTCT/PET.

\item We collect and evaluate on the largest dataset to date on chest and abdominal radiotherapy. Our dataset comprises of $651$ voxelwise-labeled OSLNs (by board-certified radiation oncologists) of $141$ esophageal cancer patients. Our system significantly improves the detection recall from $45\%$ to $67\%$ at $3$ FPs per scan, compared against the previous state-of-the-art CT-based detection method \cite{yan2019mulan}. The highest achieved recall of 0.828 for OSLNs detection is also clinically relevant and valuable.
\end{itemize}
\section{Related Work}

{\bf Generic Lesion Detection:}
There are two popular approaches for generic lesion detection: end-to-end~\cite{li2019mvp,yan20183d,yan2019mulan,zlocha2019improving} and two-stage methods \cite{ding2017accurate,ghafoorian2017deep,setio2016pulmonary,teramoto2016automated}. End-to-end methods have been extensively applied to the universal lesion detection task in the largest general lesion dataset currently available, {\it{i.e.}}, DeepLesion \cite{yan2018deeplesion}, and achieved encouraging performance. Notably, a multi-task universal lesion analysis network (MULAN) \cite{yan2019mulan} so far achieves the best detection accuracy using a 3D feature fusion strategy and Mask R-CNN~\cite{he2017mask} architecture.

In contrast, two-stage methods explicitly divide the detection task into candidate generation and FP reduction steps. The first step generates the initial candidates at a high recall and FP rate and the second step focuses on reducing the FP rate (especially the difficult ones) while maintaining a sufficient high recall. It decouples the task into easier sub-tasks and allows for the optimal design of each sub-task, which has shown to be more effective in problems like lung nodule \cite{teramoto2016automated,ding2017accurate} and brain lacune \cite{ghafoorian2017deep} detection as compared to the one-stage method. We adopt the two-stage strategy for the OSLN detection to effectively incorporate different features, {\it{i.e.}}, PET imaging, tumor distance map and high-semantic lesion attributes, into each stage. We demonstrate the necessity of our strategy by comparing with the state-of-the-art (SOTA) universal lesion detector MULAN \cite{yan2019mulan} in the experiment.

{\bf Lymph Node Detection and Segmentation:}
All previous works focus only on enlarged LN detection and segmentation in contrast-enhanced CT. Conventional statistical learning approaches   \cite{barbu2011automatic,feulner2013lymph,kitasaka2007automated,liu2016mediastinal} employ hand-crafted image features, such as shape, spatial priors, Haar filters, and volumetric directional difference filters, to capture LN appearance and location. More recent deep learning methods achieve better performance. \cite{nogues2016automatic,oda2018dense,bouget2019semantic} applies the FCN or Mask R-CNN to directly segment LNs. In contrast, \cite{roth2015improving,roth2014new} proposed a 2.5D patch-based convolutional neural network (CNN) with random view aggregation to classify LNs given all LN candidates already detected, and achieves SOTA classification accuracy for enlarged LNs. We demonstrate the effectiveness of the local and global modules in our GLNet compared with the 2.5D classification method \cite{roth2015improving}.

{\bf Multi-Modal Image Analysis:} 
The multi-modal imaging setup \cite{kuijf2019standardized,menze2014multimodal} is a common and  effective representation for segmenting anatomical structures in medical images. The pixel contrast and visual information in each modality is different and complementary for many applications. In our work, RTCT and PET have fundamentally different imaging physics, with RTCT corresponding to anatomy-based structural imaging and PET to functional imaging. Recent deep learning approaches~\cite{jin2019accurate,teramoto2016automated,xu2018automated,zhao2018tumor} have exploited different fusion strategies for PET/CT, {\it{e.g.}}, early, late or chained fusion. In our 1st-stage, we propose a $2$-stream deep network segmentation workflow (encoding RTCT alone or combined RTCT/PET and tumor distance map, respectively) and implement a concise late probability fusion scheme. This simple two-stream fusion strategy effectively generates the OSLN candidates with a high recall at a reasonable FP rate, which is  desirable for the downstream 2nd-stage FP reduction.

\begin{figure}[t]
\centering
\includegraphics[width=1.0\textwidth]{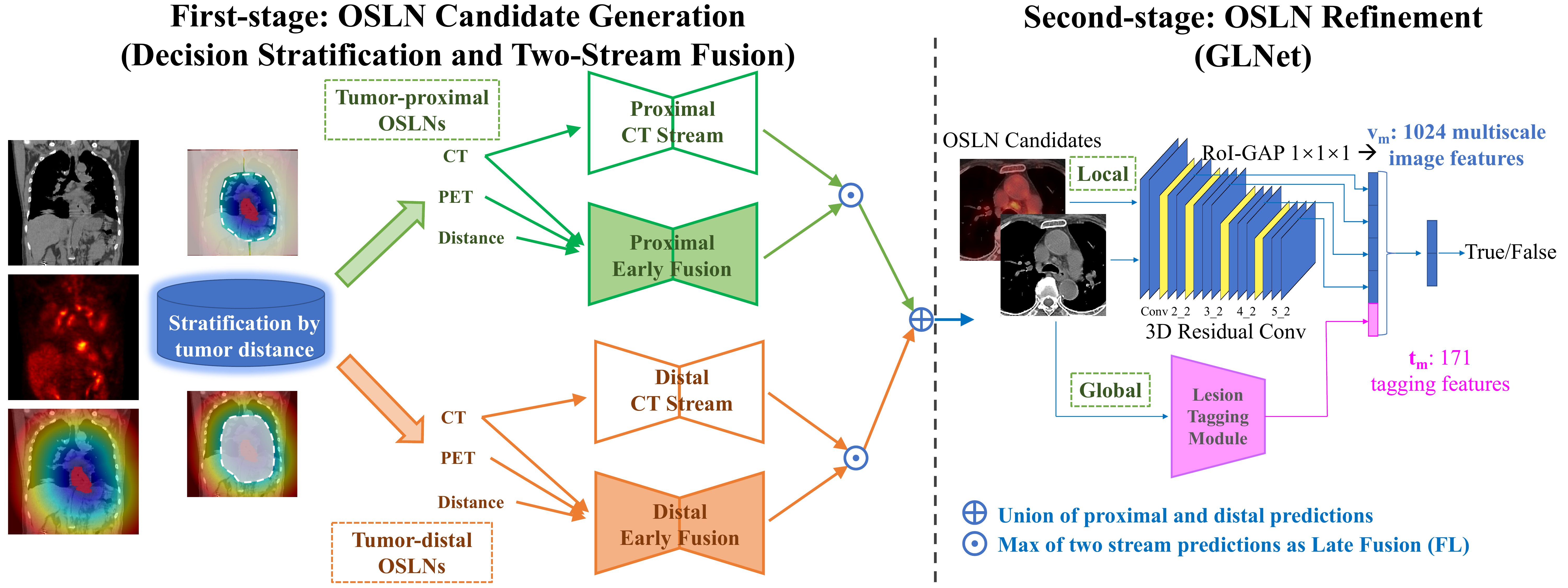}
\caption{The overall framework of our $2$-stage OSLN detection method. The 1st-stage adopts a divide-and-conquer distance stratification to divide OSLNs into tumor-proximal (green) and tumor-distal (orange) categories. For each category, a two-stream network, {\it{i.e.}}, CT stream (no fill) and CT, PET and tumor-distance early fusion stream (solid fill), is designed to learn the specific features for this category. After that, the predictions of the two streams are fused together via the ``max" operation to achieve high recall. The GLNet of the 2nd-stage takes the OSLN candidates from the 1st-stage, and passes it through the local and global modules to reject FPs, leading to a final set of OSLNs with clinically relevant recall and low FPs.}
\label{Fig:Flowchart}
\end{figure}

\section{Method}
Fig.~\ref{Fig:Flowchart} illustrates our two-stage framework, which combines OSLN candidate generation with FP rejection. In the 1st-stage, we group training-set OSLNs into two categories based on their distances to the primary tumor via distance stratification. For each category, a two-stream detection-by-segmentation network is designed to effectively incorporate and fuse the RTCT and PET images, along with a tumor distance transform map. Results from two categories are merged together to produce the OSLN candidates. The goal of the 1st-stage is to have a set of OSLN candidates with high recall while keeping FPs to a reasonable number. In the 2nd-stage, the GLNet, composed of local and global modules, is proposed to serve as a selective classifier to reject OSLN FP candidates (especially the difficult ones) while preserving sufficient recall. 

\subsection{OSLN Candidate Generation}\label{Sec:CandidateGeneration}

Assuming $N$ data samples, we denote a dataset as $\mathbf{S}=\left\{\left(\mathbf{X}^{\mathrm{CT}}_n, \mathbf{X}^{\mathrm{PET}}_n,\mathbf{Y}^{\mathrm{T}}_{n},  \mathbf{Y}^{\mathrm{LN}}_n\right)\right\}_{n=1}^{N}$, where $\mathbf{X}^{\mathrm{CT}}_n$, $\mathbf{X}^{\mathrm{PET}}_n$, $\mathbf{Y}^{\mathrm{T}}_{n}$ and $\mathbf{Y}^{\mathrm{LN}}_n$ represent the non-contrast RTCT, registered PET, the tumor mask and the ground truth LN segmentation mask, respectively.  Without loss of generality we drop $n$ for conciseness for the rest of the paper. The mask $\mathbf{Y}^{\mathrm{T}}$ is a 3D volume with a binary value $y_{i}$ at each spatial location $i$ to indicate whether the voxel $x_{i}$ is the OSLN target. 
To encode the tumor distance information, we compute the 3D signed distance transform map from the primary tumor $\mathbf{Y}^{\mathrm{T}}$, denoted as $\mathbf{X}^{\mathrm{D}}$, where each voxel $x^\mathrm{D}_i\in\mathbf{X}^{\mathrm{D}}$ represents the distance between this voxel to the nearest boundary of the primary tumor. Let $\Gamma(\mathbf{Y}^{\mathrm{T}})$ be a function that computes boundary voxels of the tumor. The distance transform value at a voxel $x^\mathrm{D}_i$ is computed as

\begin{align}
\mathbf{X}^{\mathrm{D}}(x^\mathrm{D}_i) = \left \{
\begin{array}{rcl}
\underset{q\in \Gamma(\mathbf{Y}^{\mathrm{T}})}{\min} d(x^\mathrm{D}_i,q)  & \quad {\text{if} \quad x^\mathrm{D}_i\notin \mathbf{Y}^{\mathrm{T}}}\\
-\underset{q\in \Gamma(\mathbf{Y}^{\mathrm{T}})}{\min} d(x^\mathrm{D}_i,q)  & \quad {\text{if} \quad x^\mathrm{D}_i\in \mathbf{Y}^{\mathrm{T}} }
\end{array} \right. \mathrm{,}
\end{align}
where $d(x^\mathrm{D}_i,q)$ is a distance measure from $x^\mathrm{D}_i$ to $q$. We choose to use Euclidean distance in our work and use Maurer's efficient algorithm~\cite{maurer2003linear} to compute the $\mathbf{X}^{\mathrm{D}}$. Note that $\mathbf{X}^{\mathrm{CT}}$ and $\mathbf{X}^{\mathrm{PET}}$ and $\mathbf{Y}^{\mathrm{T}}$ are already given and $\mathbf{X}^{\mathrm{D}}$ is pre-computed at the inference time.

We denote segmentation models as a mapping: $\mathbf{P} = {\mathbf{f}\!\left(\mathbf{\mathcal{X}}; \boldsymbol{\Theta}\right)}$, where $\mathbf{\mathcal{X}}$ is a set of inputs, which may consist of a single modality or a concatenation of multiple modalities. $\boldsymbol{\Theta}$ indicates model parameters, and $\mathbf{P}$ denotes the predicted probability volume. Specifically in a neural network, $\boldsymbol{\Theta}$ is parameterized by the network parameters.  
\subsubsection{Distance-Based OSLN Stratification}\label{Sec:DCS}

Based on $\mathbf{X}^{\mathrm{D}}$, we divide image voxels into two groups, $x_\mathrm{prox}$ and $x_\mathrm{dis}$, to be tumor-proximal and tumor-distal, respectively, where $\mathrm{prox}=\{i|x^\mathrm{D}_i\leq d\}$ and $\mathrm{dis}=\{i|x^\mathrm{D}_i>d\}$. In this way, we divide all OSLNs into two categories, and train separate segmentation models for each. By doing this, we break down the challenging OSLN segmentation problem into two simpler sub-problems, each of which can be more easily conquered. This allows the OSLN segmentation method to emulate the clinician decision process, where tumor-proximal LNs are more readily considered oncology-significant, whereas a more conservative process, with differing criteria, is used for tumor-distal LNs. See Fig.~\ref{Fig:Flowchart} for the distance stratification demonstration. Prediction volumes generated by the tumor-proximal or tumor-distal models are denoted as $\mathbf{P_\mathrm{prox}}$ and $\mathbf{P_\mathrm{dis}}$, respectively.

\subsubsection{Two-Stream Detection-by-Segmentation Fusion}\label{Sec:TwoSSF}
For each OSLN category, we again emulate the physician's diagnostic process by fully exploiting the complementary information within the RTCT, PET and tumor distance map. Specifically, \textit{for each OSLN category}, we design a two-stream 3D segmentation workflow that fuses predictions from two independent sub-networks, one trained using the RTCT alone (\textit{CT stream}), and the other trained using the three channels of RTCT, PET and the tumor distance map jointly (\textit{early fusion stream}). In this way we generate predictions based on only structural appearance, complementing them with additional predictions incorporating PET's auxiliary functional sensitivity and the tumor distance-map's location context. We denote prediction volumes from the RTCT and early fusion stream models  as  $\mathbf{P^{\mathrm{CT}}_{(.)}}$ and $\mathbf{P^{\mathrm{EF}}_{(.)}}$, respectively, where the subscript may be either ``$\mathrm{prox}$'' or ``$\mathrm{dis}$'' for the tumor-proximal or tumor-distal categories, respectively. The result is four separate predictions. To ensure a high recall of OSLN detection in this stage, we apply a straightforward yet effective \textit{late fusion} by taking the element-wise $\texttt{max}$ and $\texttt{union}$ operations of the four predictions:
\begin{equation}\label{Eq:DCSModel}
 \mathbf{P^{LF}}=\{p_{i}|p_i=\texttt{union}\{ \texttt{max}\{p^{\mathrm{CT}}_{\mathrm{prox},i},p^{\mathrm{EF}}_{\mathrm{prox},i}\}, \texttt{max}\{p^{\mathrm{CT}}_{\mathrm{dis},i},p^{\mathrm{EF}}_{\mathrm{dis},i}\}\}\},
\end{equation}
where ${p^{(.)}_{(.),i}}\in \mathbf{P^{(.)}_\mathrm{(.)}}$ and $i$ indexes individual voxel locations. Stratifying OSLNs by tumor distance and performing two stream fusion are both crucial for a high recall.

From the final segmentation probability $\mathbf{P^{LF}}$, we derive the binary segmentation mask $\mathbf{B}$ by thresholding, and then calculate the OSLN instance candidates as the input to the $2$nd-stage.

\subsection{OSLN Refinement by Global-Local Classification}\label{Sec:FPReudction}

The goal of the 2nd-stage is to reject as many FPs as possible while maintaining a sufficiently high recall. We first aggregate all predicted OSLN instances from the 1st-stage to be $\mathbf{R}=\left\{\left(\mathbf{C}^{\mathrm{CT}}_m, \mathbf{C}^{\mathrm{PET}}_m, l_m\right)\right\}_{m=1}^{M}$ as the OSLN candidates set, where $\mathbf{C}^\mathrm{CT}_m$ and $\mathbf{C}^\mathrm{PET}_m$ denote the local RTCT and PET image patches cropped at the $m$th OSLN candidate, respectively, and the binary scalar $l_m$ is the label indicating if this instance is a true OSLN. We formulate a classification model: $q = {\mathbf{g}\!\left(\mathbf{C}; \boldsymbol{\Phi}\right)}$, where $\mathbf{C}$ represents the input image patches, $\boldsymbol{\Phi}$ stands for model parameters, and $q$ denotes the predicted probability. Here, when appropriate, we drop the $m$ for simplicity.  

To design a highly effective OSLN classifier, especially for the hard FPs, we propose a global and local network (GLNet) to leverage both local (CT appearance and PET signals) and global (spatial prior and other attributes) features. We describe their details in the following subsections.
\subsubsection{Local module in GLNet}
For the local module, we adopt a multi-scale 3D CNN model with a 3D ROI-GAP pooling layer~\cite{girshick2015fast} to extract OSLN local features from the image patch $\mathbf{C}$. Unlike the 2.5D input patch used in~\cite{roth2015improving}, the 3D CNN explicitly uses 3D spatial information, improving classification performance. Either {CT} or {CT}+{PET} patches can be fed into the local model, and we evaluate both options. The features generated by each convolutional block separately pass through a 3D ROI-GAP pooling layer and a fully connected layer to form a $256$D vector, which are then concatenated together to a multi-scale local representation for the OSLN instance. Since we use four CNN blocks, this leads to a total of $4\times256=1024$-dimensional feature vector, which is denoted as $\mathbf{v}$. See the 2nd-stage illustration in Fig.~\ref{Fig:Flowchart}.

\subsubsection{Global module in GLNet}
For the global module, we migrate the ontology-based medical knowledge from the large-scale DeepLesion~\cite{yan2018deeplesion} dataset, via a pretrained lesion tagging module, {\it{i.e.}}, LesaNet~\cite{yan2019holistic}. Trained from radiology reports, LesaNet predicts high-level semantic lesion properties in the form of a $171$-dimensional vector describing the lesion's body part, type and attributes. These information may not be easily captured from local image patches. We use the prediction of LesaNet on the $m$th OSLN candidates to generate a $171$-dimensional feature vector $\mathbf{t}_m$, which provides complementary information to distinguish a true OSLN from false ones. For example, one body-part attribute from the feature vector indicates whether the lesion is in the ``muscle", which may be confused with OSLNs when only analyzing the small local image patch, but are easier to identify under a global context. These kinds of FP candidates can be safely rejected using the global properties. LesaNet also predicts body parts like hilum LN, subcarinal LN, pretracheal LN and attributes like hypo-attenuation, tiny, oval, which are all relevant properties to distinguish true OSLNs from false ones.

To combine the strength of local image-based features and global OSLN properties, the GLNet concatenates $\mathbf{v}_m$ and $\mathbf{t}_m$ and passes through a fully connected layer to generate the final OSLN classification score, as illustrated in Fig.~\ref{Fig:Flowchart}. 
\section{Experiments}

\begin{figure}[t]
    \centering
    \includegraphics[width=1.0\linewidth]{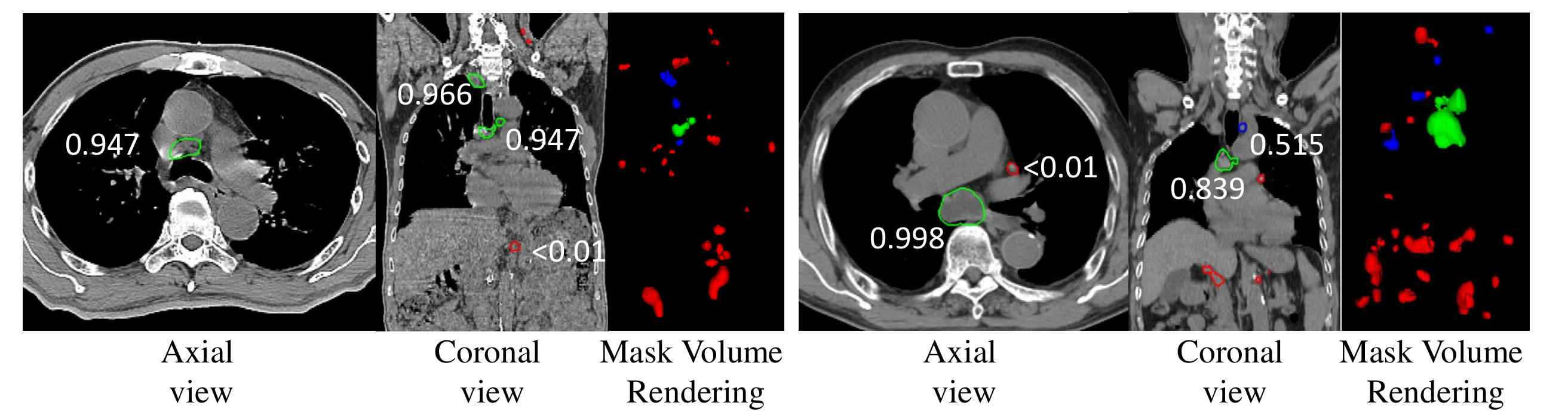}
    \caption{Visualization of segmentation contours in axial view or coronal views, and 3D mask volume rendering of two cases (left, and right). All masks/contours are LNs candidates from the first stage, where red ones are rejected in the 2nd-stage. Compared with ground truth LNs, TP  and FP are colored in green and blue, respectively. Best viewed in color.}
    \label{Fig:LNRejected}
    \vspace{-5mm}
\end{figure}

In Fig.~\ref{Fig:LNRejected}, we provide visual examples of our OSLN detection results. 
First, we can find that a large number of OSLN candidates are generated after the $1$st-stage, to warrant a high recall. Second, the majority of OSLN candidates are effectively reduced by our proposed GLNet classifier, while the true positives in the 1st-stage are kept after the false positive reduction, which is desirable. Below we elaborate further on our experiments, providing dataset and implementation details along with extensive quantitative analyses. 

\subsection{Datasets}
We collected an in-house dataset to evaluate our 1st-stage performance as well as the overall two-stage performance. We collected $141$ non-contrast RTCTs of anonymized esophageal cancer patients, all undergoing radiotherapy treatments. Radiation oncologists labeled the 3D segmentation masks of the primary tumor and all OSLNs treated by radiotherapy. In total, there is a non-contrast RTCT scan and a PET/CT for each of the $141$ patients and $651$ OSLNs with voxel-wise labels in the mediastinum or upper abdomen regions. This is {\it{the largest annotated OSLN dataset}} in the chest and abdominal region to-date. We register the PET images to RTCT using the registration method in~\cite{jin2019accurate}. For evaluation, we randomly split the annotated $141$ patients into $84$ for training, $23$ for validation, and $34$ for testing. In our experiments, we resample RTCT and PET images to have a consistent spatial resolution of $1\times1\times2.5$ mm. For data preprocessing, we truncate Hounsfield unit values of the RTCT to be within $[-200, 300]$. We also calculate the mean and standard deviation values of PET images across the entire training set and then normalize all PET images with these values.
\subsection{Implementation Details}
In the first stage, for training the OSLN detection-by-segmentation network, we crop sub-volumes of $96\times96\times64$ from the 3D images of RTCT, registered PET and the tumor-LN distance map. For the distance stratification, we set $d=70$ mm to divide OSLN instances to tumor-proximal and tumor-distal sub-groups as suggested by our physician, and train the tumor-proximal and tumor-distal models separately. For data augmentation, we use straightforward and effective augmentations on training patches, {\it{i.e.}}, rotation ($90\degree$, $180\degree$, and $270\degree$) with a probability of $0.5$ and flips in the axial view with a probability of $0.25$. We can choose any popular segmentation network as our 1st-stage backbone, and we opt for the standard 3D UNet~\cite{cciccek20163d} as it gives the best performance in our network backbone ablation study in Sec.~\ref{Sec:ablation_1}. Models are trained on two NVIDIA Quadro RTX 6000 GPUs with a batch size of $8$ for $50$ epochs. The RAdam~\cite{liu2019variance} optimizer with a learning rate of $0.0001$ is used with a momentum of $0.9$ and a weight decay of $0.0005$. For testing, we use a computationally efficient way to inference, {\it{i.e.}}, sub-volumes of $224\times224\times64$ are cropped along the vertical axis with the horizontal center the same as the center of lung masks~\cite{harrison2017progressive}. These sub-volume predictions are aggregated to obtain the final OSLN segmentation results.

In the 2nd-stage, to train the local module of GLNet, the input images are generated by cropping a $48 \times 48 \times 32$ sub-volume centered around each predicted OSLN candidate from the 1st-stage. If the size of the predicted OSLN is larger than $48 \times 48 \times 32$, we resize the sub-volume so that it contains at least an $8$-voxel margin of the background along each dimension to ensure sufficient background context. The bounding boxes (bbox) for the 3D ROI-GAP pooling layer in Sec.~\ref{Sec:FPReudction} are generated by randomly jittering the bbox around the predicted OSLN with a $3$-voxel range in each dimension. For the global module of GLNet, we use the publicly available LesaNet~\cite{yan2019holistic} pre-trained on the DeepLesion dataset. The input of LesaNet is a 120$\times$120 2D CT image patch around the OSLN candidate. The overall GLNet is trained using Adam~\cite{kingma2014adam} optimizer with a learning rate of $0.0001$ and batch size of $32$ for $10$ epochs. 

\subsection{Evaluation Metrics}\label{Sec:Eva}
We first describe the hit, {\it{i.e.}}, the correct detection, criteria for OSLN detection when using the segmentation results. For an OSLN prediction from the $1$st-stage, if it overlaps with any ground-truth OSLN, we treat it as a hit provided that its estimated radius is similar to the radius of the ground-truth OSLN. After confirming with our physician, a predicted radius must be within a factor of $[0.5, 1.5]$ to the ground-truth radius.

\subsubsection{Recall and Precision}\label{Sec:EvaRecall}
We assess the performance of the 1st-stage by reporting the recall at a range of desired precision points. Note that the goal of the 1st-stage is to achieve a high recall (even with quite a few FPs) so that the 2nd-stage has a high upper-bound recall to work with while it filters out FPs. We report the mean recall (mRecall) at a precision range of $[0.10, 0.20]$ to reflect the model performance. We also report the recall at a precision of $0.15$, which is the operating point we choose to generate inputs for the 2nd-stage. This operating point was chosen after confirming with our radiation oncologist. Both the recall and precision are macro-averaged across patients. 

\subsubsection{FROC}
To evaluate both the complete workflow (1st+2nd-stage), we compute the free response operating characteristic (FROC), which measures the recall against different numbers of FPs allowed per patient. We report the average recall (mFROC) at $2$, $3$, $4$, $6$ FPs per patient study. Besides the mFROC, we also report the best F$1$ score a model can achieve. 
     \begin{table}[t]
     \caption{Ablation study on the validation set} 
     \begin{subtable}{0.5\textwidth}
     \caption{Ablation study of different backbones for the CT and early fusion streams} %
     \begin{tabular}{lccrl}\toprule
          Backbone &\multicolumn{2}{c}{Recall@{$0.15$}}&\multicolumn{2}{c}{mRecall@{$0.10\text{-}0.20$}}\\\cmidrule(lr){2-3}\cmidrule(lr){4-5}
           &{CT}& {EF}&{CT}& {EF} \\
    \hline
           3D-UNet &\bm{$0.736$} &\bm{$0.732$} &\bm{$0.762$} &\bm{$0.722$} \\
           SE-UNet &$0.686$ &$0.705$ &$0.693$ &$0.705$\\
           HRNet &$0.524$ &$0.656$ &$0.538$ &$0.638$\\
           PSNN &$0.709$ &$0.574$ &$0.714$ &$0.592$\\
     \bottomrule
     \end{tabular}
     \label{Tab:AblationBackbone}
     \end{subtable}
     \begin{subtable}{0.5\textwidth}
     \caption{3D UNet performance with (``w/'') and without(``w/o'') distance stratfication. All three settings, CT, EF, and LF, are tested
      }
      \begin{tabular}{lcccc}\toprule
          Input &\multicolumn{2}{c}{Recall@{$0.15$}}&\multicolumn{2}{c}{mRecall@{$0.10\text{-}0.20$}}\\\cmidrule(lr){2-3}\cmidrule(lr){4-5}
           &{w/}& {w/o}&{w/}& {w/o} \\
    \hline
           LF &\bm{$0.828$} &$\bm{0.786}$ &\bm{$0.817$} &$0.732$\\
    \hline
           EF &$0.788$ &$0.732$ &$0.760$ &$0.722$\\
           CT &$0.772$ &$0.736$ &$0.772$ &$\bm{0.762}$\\
     \bottomrule
     \end{tabular}
     \label{Tab:AblationStratification}
     \end{subtable}
     \vspace{-5mm}
 \end{table}

\subsection{1st-Stage Ablation Study} \label{Sec:ablation_1}

\subsubsection{Segmentation Network Backbone} 
We evaluated different segmentation backbones for the OSLN candidate generation, {\it{i.e.}}, standard UNet~\cite{cciccek20163d}, UNet with squeeze-and-excitation (SE) block~\cite{hu2018squeeze}, HRNet~\cite{sun2019deep}, and PSNN~\cite{jin2019accurate}. As shown in Table~\ref{Tab:AblationBackbone}, the standard 3D UNet~\cite{cciccek20163d} consistently outperforms other backbones. For PSNN~\cite{jin2019accurate}, it probably has difficulty handling this challenging task (dealing with small objects) due to its simplistic ``upsampling" decoders. For the HRNet~\cite{sun2019deep}, due to its memory-hungry computations, we can only add the high resolution features after two pooling layers, which is undesired for segmenting OSLNs. The attention module from the SE block~\cite{hu2018squeeze} does not help with this segmentation task either.

\vspace{-2.5mm}
\subsubsection{Distance stratification and Two-Stream Network Fusion}
We verify the effectiveness of the proposed distance stratification method under different settings. As shown in Table.~\ref{Tab:AblationStratification}, among all settings, {\it{i.e.}}, CT, early fusion (EF), and late fusion (LF), the distance stratification consistently improves recall@$0.15$ by $4\%-5\%$. Similar improvements are seen for mRecall@$0.1$-$0.2$. These results strongly support our use of distance stratification, which is shown to be effective under different input settings. 

Table~\ref{Tab:AblationStratification} also reveals the importance of using and fusing different streams. As we can see, the CT stream and the EF stream achieve similar performance to each other, regardless of whether distance stratification is used or not. However, when the two streams are combined together using LF, marked improvements are observed. For example, the recall@$0.15$ gains $4\%$-$5\%$, and the mRecall@$0.1$-$0.2$ shows similar improvements. These quantitative results validate the effectiveness of the proposed distance stratification and the two-stream network fusion.

\subsection{2nd-stage Ablation Study}

\subsubsection{Necessity of the 2nd-stage}
To gauge the impact of the 2nd-stage, we first directly evaluate the OSLN detection accuracy using the 1st-stage alone. Specifically, the detection score of each OSLN instance is determined by averaging the segmentation probability for every voxel within the segmentation mask. All ``1st-stage only'' results in Tab.~\ref{Tab:FullResults} are marked by ``$\#$". Focusing first on the LF setting, when using the 1st-stage alone it provides $0.441$ F$1$ and $0.478$ mFROC. When adding a second-stage classifier only accepting CT as input, the F$1$ scores and mFROC are improved to $0.513$ and $0.576$, respectively. Providing the PET image and global tags to the 2nd-stage classifier boosts performance even further to $0.552$ and $0.645$ for F$1$ scores and mFROC, respectively. These are clinically impactful gains.  Finally, regardless of the 1st-stage setting (LF, EF, or CT), the 2nd-stage classifier provides clear improvement. This proves the versatility and strength of our workflow. 

\subsubsection{Role of Local and Global Modules in GLNet}
To show the necessity of both the local and global GLNet modules, we also evaluated purely local and purely global 2nd-stage classification performance. As can be seen in Table~\ref{Tab:FullResults}, regardless of which 1st-stage setting is used, a purely local 2nd-stage (\eg{} last 2nd and 3rd rows) outperforms a purely global 2nd-stage (\eg{} last 4th row). This indicates that the high-level semantic features migrated from the general lesion tagging model, {\it{i.e.}}, LesaNet~\cite{yan2019holistic}, are less effective than the local OSLN features extracted from CT or CT+PET. However, when combining the global tags with the local patches using the proposed GLNet, mFROC performance is increased from $0.594$ to $0.645$ (when using the LF 1st-stage setting). This demonstrates that both local and global features contribute to our ultimate performance. These observations are also valid when using the CT or EF settings for the 1st-stage. 

\begin{table}[t]
\caption{Performance comparison of different methods on the testing set. The ``1st-Stage Setting" column denotes which setting is used to generate OSLN candidates. ``\#" means we directly evaluate based on 1st-stage instance-wise segmentation scores. The ``2nd-Stage Inputs" column indicates which inputs are provided to the 2nd-stage classifier. Boldface denotes our chosen 2nd-stage classifier, evaluated across different 1st-stage settings. We also compare against previous state-of-the-arts, the~\cite{roth2015improving} and the end-to-end MULAN system~\cite{yan2019mulan}}
\centering
\begin{small}
\begin{tabular}{lccc|cc}\toprule
                           1st-Stage Setting
                           &\multicolumn{3}{c}{2nd-Stage Inputs\hphantom{.3in}} &\multicolumn{2}{c}{Evaluation Metrics} \\\cmidrule(lr){2-4}\cmidrule(lr){5-6} &CT & PET &{Tag}& {F$1$}&{mFROC} \\ \hline

{CT\#} &   &   &                                      & 0.407 & 0.431 \\ %
{EF\#} &    &  Not Applied  &                                     & 0.370 & 0.395 \\ %
{LF\#} &    &    &                                & 0.441 & 0.478 \\ %
\hline
{CT}~\cite{roth2015improving} & \checkmark & &       & 0.220 & 0.067 \\ %
{CT} &   &   & \checkmark                            & 0.380 & 0.408 \\ %
{CT} & \checkmark  &  &                              & 0.421 & 0.449 \\ %
{CT} & \checkmark  & \checkmark  &                   & 0.450 & 0.491 \\ %
{{\bf{CT (GLNet)}}} & \checkmark & \checkmark  &\checkmark          & 0.513 & 0.563 \\ %
\hline
{EF~\cite{roth2015improving}} & \checkmark&    &    & 0.225 & 0.092 \\ %
{EF} &   &   & \checkmark                           & 0.397 & 0.444 \\ %
{EF} & \checkmark  &  &                             & 0.423 & 0.473 \\ %
{EF} & \checkmark  & \checkmark  &                  & 0.469 & 0.518 \\ %
{{\bf{EF (GLNet)}}} & \checkmark & \checkmark  &\checkmark         & 0.507 & 0.572 \\ %
\hline
{LF~\cite{roth2015improving}} & \checkmark &&    & 0.257 & 0.143 \\ %
{LF} &   &   & \checkmark                        & 0.471 & 0.531 \\ 
{LF} & \checkmark  &  &                          & 0.513 & 0.576 \\ %
{LF} & \checkmark  & \checkmark  &               & 0.526 & 0.594 \\ %
{{\bf{LF (GLNet)}}} & \checkmark & \checkmark  &\checkmark      & $\bm{0.552}$ & $\bm{0.645}$ \\%
\toprule
End-to-End Method\hphantom{.3in}
                           &\multicolumn{3}{c}{Inputs} &\multicolumn{2}{c}{Evaluation Metrics} \\\cmidrule(lr){2-4}\cmidrule(lr){5-6} &CT & PET &{Tag}& {F$1$}&{mFROC} \\ \hline
{MULAN~\cite{yan2019mulan}} & \checkmark      & \checkmark      &  \checkmark & 0.436 & 0.475 \\
{MULAN~\cite{yan2019mulan}} & \checkmark      &            &    \checkmark    & 0.335 & 0.348 \\
\bottomrule
\end{tabular}
\end{small}
\vspace{-6mm}
\label{Tab:FullResults}
\end{table}

\subsection{Comparison to the State-of-the-Art}
Table.~\ref{Tab:FullResults} also compares the proposed two-stage OSLN detection method with 2 state-of-the-art methods, {\it{i.e.}}, the multi-task universal lesion analysis network (MULAN)~\cite{yan2019mulan} (achieves the best general lesion detection results in the DeepLesion dataset) and a 2.5D CNN method for classifying enlarged LNs~\cite{roth2015improving} (achieves the best 2nd-stage LN classification results in the enlarged LN dataset). We retrain the MULAN using both CT and CT+PET as inputs on our radiotherapy dataset. The tagging information is naturally incorporated in MULAN regardless of input channels. Several conclusions can be drawn. First, MULAN's results, based on the CT+PET input (0.475 mFROC), are better than those based on the CT alone (0.348 mFROC), which again demonstrates the importance of PET imaging in the OSLN detecting task, even when using a single end-to-end trained model. Second, MULAN's best performance is just comparable with our best 1st-stage-only results, {\it{i.e.}}, (LF$\#$). This demonstrates the effectiveness of our 1st-stage with distance stratification and the two-stream network fusion. Third, our complete pipeline, regardless of the 1st-stage settings, significantly outperforms the best MULAN results, {\it{e.g.}}, CT (GLNet) achieves an mFROC score of $0.563$ as compared to $0.475$ from MULAN, whereas LF (GLNet) further boosts the mFROC to $0.645$. This is a $22\%$ improvement and highlights the advantages of our two-stage method, which is tailored to achieve maximum performance gain on the challenging and unique OSLN problem. 

Similar to our 2nd-stage, the 2.5D CNN method of~\cite{roth2015improving} is designed to classify LN candidates, but it was characterized only on enlarged LN candidates using contrast-enhanced CT. We trained it using our non-contrast CT local patches under different 1st-stage settings, {\it{i.e.}}, CT, EF and LF. Note that it has the worst performance among all 2nd-stage classifiers, with a best mFROC of only 0.143. This large performance degradation, particularly compared to our CT-only 2nd-stage classifier, is probably due to its 2.5D input setup and the missing of PET information. Although the 2.5D inputs and 3 orthogonal views is efficient for enlarged LN classification~\cite{roth2015improving}, this pseudo 3D analysis cannot fully leverage the 3D information that seems important to differentiate OSLNs from background. 

\section{Conclusion}
We proposed a new two-stage approach to automatically detect and segment oncology significant lymph nodes (OSLNs) from non-contrast CT and PET, which has not
been previously studied as a computational task. In the 1st-stage, we introduce a divide-and-conquer distance stratification method by dividing OSLNs into tumor-proximal and tumor-distal categories; followed by training separate detection-by-segmentation networks to learn the category specific features aimed to decouple this challenging task into two easier ones. In the 2nd-stage, we propose the GLNet to further reduce the false positives from the 1st-stage, by combining local appearance features from CT/PET patches and global semantic information migrated from a general lesion-characteristics-tagging model. Our method is evaluated on the largest OSLN dataset of $141$ esophageal cancer patients. Our proposed framework significantly improves the recall from $45\%$ to $67\%$ at the $3$ false-positive rates per patient as compared to previous state-of-the-art methods. Thus, our work represents an important step forward toward OSLNs detection and segmentation. 

\clearpage
%
%
\bibliographystyle{splncs04}
\bibliography{egbib}

\begin{thebibliography}{10}
\providecommand{\url}[1]{\texttt{#1}}
\providecommand{\urlprefix}{URL }
\providecommand{\doi}[1]{https://doi.org/#1}

\bibitem{barbu2011automatic}
Barbu, A., Suehling, M., Xu, X., Liu, D., Zhou, S.K., Comaniciu, D.: Automatic
  detection and segmentation of lymph nodes from ct data. IEEE Transactions on
  Medical Imaging  \textbf{31}(2),  240--250 (2011)

\bibitem{bouget2019semantic}
Bouget, D., J{\o}rgensen, A., Kiss, G., Leira, H.O., Lang{\o}, T.: Semantic
  segmentation and detection of mediastinal lymph nodes and anatomical
  structures in ct data for lung cancer staging. International journal of
  computer assisted radiology and surgery pp. 1--10 (2019)

\bibitem{cciccek20163d}
{\c{C}}i{\c{c}}ek, {\"O}., Abdulkadir, A., Lienkamp, S.S., Brox, T.,
  Ronneberger, O.: {3D} u-net: learning dense volumetric segmentation from
  sparse annotation. In: MICCAI (2016)

\bibitem{ding2017accurate}
Ding, J., Li, A., Hu, Z., Wang, L.: Accurate pulmonary nodule detection in
  computed tomography images using deep convolutional neural networks. In:
  International Conference on Medical Image Computing and Computer-Assisted
  Intervention. pp. 559--567. Springer (2017)

\bibitem{eisenhauer2009new}
Eisenhauer, E.A., Therasse, P., Bogaerts, J., Schwartz, L.H., Sargent, D.,
  Ford, R., Dancey, J., Arbuck, S., Gwyther, S., Mooney, M., et~al.: New
  response evaluation criteria in solid tumours: revised recist guideline
  (version 1.1). European journal of cancer  \textbf{45}(2),  228--247 (2009)

\bibitem{feulner2013lymph}
Feulner, J., Zhou, S.K., Hammon, M., Hornegger, J., Comaniciu, D.: Lymph node
  detection and segmentation in chest ct data using discriminative learning and
  a spatial prior. Medical image analysis  \textbf{17}(2),  254--270 (2013)

\bibitem{ghafoorian2017deep}
Ghafoorian, M., Karssemeijer, N., Heskes, T., Bergkamp, M., Wissink, J., Obels,
  J., Keizer, K., de~Leeuw, F.E., van Ginneken, B., Marchiori, E., et~al.: Deep
  multi-scale location-aware 3d convolutional neural networks for automated
  detection of lacunes of presumed vascular origin. NeuroImage: Clinical
  \textbf{14},  391--399 (2017)

\bibitem{girshick2015fast}
Girshick, R.: Fast r-cnn. In: Proceedings of the IEEE international conference
  on computer vision. pp. 1440--1448 (2015)

\bibitem{goel2017clinical}
Goel, R., Moore, W., Sumer, B., Khan, S., Sher, D., Subramaniam, R.M.: Clinical
  practice in pet/ct for the management of head and neck squamous cell cancer.
  American Journal of Roentgenology  \textbf{209}(2),  289--303 (2017)

\bibitem{harrison2017progressive}
Harrison, A.P., Xu, Z., George, K., Lu, L., Summers, R.M., Mollura, D.J.:
  Progressive and multi-path holistically nested neural networks for
  pathological lung segmentation from ct images. In: International conference
  on medical image computing and computer-assisted intervention. pp. 621--629.
  Springer (2017)

\bibitem{he2017mask}
He, K., Gkioxari, G., Doll{\'a}r, P., Girshick, R.: Mask r-cnn. In: Proceedings
  of the IEEE international conference on computer vision. pp. 2961--2969
  (2017)

\bibitem{he2016deep}
He, K., Zhang, X., Ren, S., Sun, J.: Deep residual learning for image
  recognition. In: Proceedings of the IEEE conference on computer vision and
  pattern recognition. pp. 770--778 (2016)

\bibitem{hu2018squeeze}
Hu, J., Shen, L., Sun, G.: Squeeze-and-excitation networks. In: Proceedings of
  the IEEE conference on computer vision and pattern recognition. pp.
  7132--7141 (2018)

\bibitem{jin2019accurate}
Jin, D., Guo, D., Ho, T.Y., Harrison, A.P., Xiao, J., Tseng, C.k., Lu, L.:
  Accurate esophageal gross tumor volume segmentation in pet/ct using
  two-stream chained 3d deep network fusion. In: International Conference on
  Medical Image Computing and Computer-Assisted Intervention. pp. 182--191.
  Springer (2019)

\bibitem{kingma2014adam}
Kingma, D.P., Ba, J.: Adam: A method for stochastic optimization. arXiv
  preprint arXiv:1412.6980  (2014)

\bibitem{kitasaka2007automated}
Kitasaka, T., Tsujimura, Y., Nakamura, Y., Mori, K., Suenaga, Y., Ito, M.,
  Nawano, S.: Automated extraction of lymph nodes from 3-d abdominal ct images
  using 3-d minimum directional difference filter. In: International Conference
  on Medical Image Computing and Computer-Assisted Intervention. pp. 336--343.
  Springer (2007)

\bibitem{kuijf2019standardized}
Kuijf, H.J., Biesbroek, J.M., de~Bresser, J., Heinen, R., Andermatt, S., Bento,
  M., Berseth, M., Belyaev, M., Cardoso, M.J., Casamitjana, A., et~al.:
  Standardized assessment of automatic segmentation of white matter
  hyperintensities; results of the wmh segmentation challenge. IEEE
  transactions on medical imaging  (2019)

\bibitem{leong2006prospective}
Leong, T., Everitt, C., Yuen, K., Condron, S., Hui, A., Ngan, S.Y., Pitman, A.,
  Lau, E.W., MacManus, M., Binns, D., et~al.: A prospective study to evaluate
  the impact of fdg-pet on ct-based radiotherapy treatment planning for
  oesophageal cancer. Radiotherapy and oncology  \textbf{78}(3),  254--261
  (2006)

\bibitem{li2019mvp}
Li, Z., Zhang, S., Zhang, J., Huang, K., Wang, Y., Yu, Y.: Mvp-net: Multi-view
  fpn with position-aware attention for deep universal lesion detection. In:
  International Conference on Medical Image Computing and Computer-Assisted
  Intervention. pp. 13--21. Springer (2019)

\bibitem{liu2016mediastinal}
Liu, J., Hoffman, J., Zhao, J., Yao, J., Lu, L., Kim, L., Turkbey, E.B.,
  Summers, R.M.: Mediastinal lymph node detection and station mapping on chest
  ct using spatial priors and random forest. Medical physics  \textbf{43}(7),
  4362--4374 (2016)

\bibitem{liu2019variance}
Liu, L., Jiang, H., He, P., Chen, W., Liu, X., Gao, J., Han, J.: On the
  variance of the adaptive learning rate and beyond. arXiv preprint
  arXiv:1908.03265  (2019)

\bibitem{maurer2003linear}
Maurer, C.R., Qi, R., Raghavan, V.: A linear time algorithm for computing exact
  euclidean distance transforms of binary images in arbitrary dimensions. IEEE
  Transactions on Pattern Analysis and Machine Intelligence  \textbf{25}(2),
  265--270 (2003)

\bibitem{menze2014multimodal}
Menze, B.H., Jakab, A., Bauer, S., Kalpathy-Cramer, J., Farahani, K., Kirby,
  J., Burren, Y., Porz, N., Slotboom, J., Wiest, R., et~al.: The multimodal
  brain tumor image segmentation benchmark (brats). IEEE transactions on
  medical imaging  \textbf{34}(10),  1993--2024 (2014)

\bibitem{National2020}
Network, N.C.C.: Nccn clinical practice guidelines:head and neck cancers.
  American Journal of Roentgenology  \textbf{version 2} (2020)

\bibitem{nogues2016automatic}
Nogues, I., Lu, L., Wang, X., Roth, H., Bertasius, G., Lay, N., Shi, J.,
  Tsehay, Y., Summers, R.M.: Automatic lymph node cluster segmentation using
  holistically-nested neural networks and structured optimization in ct images.
  In: International Conference on Medical Image Computing and Computer-Assisted
  Intervention. pp. 388--397. Springer (2016)

\bibitem{oda2018dense}
Oda, H., Roth, H.R., Bhatia, K.K., Oda, M., Kitasaka, T., Iwano, S., Homma, H.,
  Takabatake, H., Mori, M., Natori, H., et~al.: Dense volumetric detection and
  segmentation of mediastinal lymph nodes in chest ct images. In: Medical
  Imaging 2018: Computer-Aided Diagnosis. vol. 10575, p. 1057502. International
  Society for Optics and Photonics (2018)

\bibitem{roth2015improving}
Roth, H.R., Lu, L., Liu, J., Yao, J., Seff, A., Cherry, K., Kim, L., Summers,
  R.M.: Improving computer-aided detection using convolutional neural networks
  and random view aggregation. IEEE transactions on medical imaging
  \textbf{35}(5),  1170--1181 (2016)

\bibitem{roth2014new}
Roth, H.R., Lu, L., Seff, A., Cherry, K.M., Hoffman, J., Wang, S., Liu, J.,
  Turkbey, E., Summers, R.M.: A new 2.5 d representation for lymph node
  detection using random sets of deep convolutional neural network
  observations. In: International conference on medical image computing and
  computer-assisted intervention. pp. 520--527. Springer (2014)

\bibitem{scatarige1983low}
Scatarige, J.C., Fishman, E.K., Kuhajda, F.P., Taylor, G.A., Siegelman, S.S.:
  Low attenuation nodal metastases in testicular carcinoma. Journal of computer
  assisted tomography  \textbf{7}(4),  682--687 (1983)

\bibitem{schwartz2009evaluation}
Schwartz, L., Bogaerts, J., Ford, R., Shankar, L., Therasse, P., Gwyther, S.,
  Eisenhauer, E.: Evaluation of lymph nodes with recist 1.1. European journal
  of cancer  \textbf{45}(2),  261--267 (2009)

\bibitem{setio2016pulmonary}
Setio, A.A.A., Ciompi, F., Litjens, G., Gerke, P., Jacobs, C., Van~Riel, S.J.,
  Wille, M.M.W., Naqibullah, M., S{\'a}nchez, C.I., van Ginneken, B.: Pulmonary
  nodule detection in ct images: false positive reduction using multi-view
  convolutional networks. IEEE transactions on medical imaging  \textbf{35}(5),
   1160--1169 (2016)

\bibitem{sun2019deep}
Sun, K., Xiao, B., Liu, D., Wang, J.: Deep high-resolution representation
  learning for human pose estimation (2019)

\bibitem{teramoto2016automated}
Teramoto, A., Fujita, H., Yamamuro, O., Tamaki, T.: Automated detection of
  pulmonary nodules in pet/ct images: Ensemble false-positive reduction using a
  convolutional neural network technique. Medical physics  \textbf{43}(6Part1),
   2821--2827 (2016)

\bibitem{xu2018automated}
Xu, L., Tetteh, G., Lipkova, J., Zhao, Y., Li, H., Christ, P., Piraud, M.,
  Buck, A., Shi, K., Menze, B.H.: Automated whole-body bone lesion detection
  for multiple myeloma on 68ga-pentixafor pet/ct imaging using deep learning
  methods. Contrast media \& molecular imaging  \textbf{2018} (2018)

\bibitem{yan20183d}
Yan, K., Bagheri, M., Summers, R.M.: 3d context enhanced region-based
  convolutional neural network for end-to-end lesion detection. In:
  International Conference on Medical Image Computing and Computer-Assisted
  Intervention. pp. 511--519. Springer (2018)

\bibitem{yan2019holistic}
Yan, K., Peng, Y., Sandfort, V., Bagheri, M., Lu, Z., Summers, R.M.: Holistic
  and comprehensive annotation of clinically significant findings on diverse ct
  images: Learning from radiology reports and label ontology. In: Proceedings
  of the IEEE Conference on Computer Vision and Pattern Recognition. pp.
  8523--8532 (2019)

\bibitem{yan2019mulan}
Yan, K., Tang, Y., Peng, Y., Sandfort, V., Bagheri, M., Lu, Z., Summers, R.M.:
  Mulan: Multitask universal lesion analysis network for joint lesion
  detection, tagging, and segmentation. In: International Conference on Medical
  Image Computing and Computer-Assisted Intervention. pp. 194--202. Springer
  (2019)

\bibitem{yan2018deeplesion}
Yan, K., Wang, X., Lu, L., Summers, R.M.: Deeplesion: automated mining of
  large-scale lesion annotations and universal lesion detection with deep
  learning. Journal of Medical Imaging  \textbf{5}(3),  036501 (2018)

\bibitem{zhao2018tumor}
Zhao, X., Li, L., Lu, W., Tan, S.: Tumor co-segmentation in pet/ct using
  multi-modality fully convolutional neural network. Physics in Medicine \&
  Biology  \textbf{64}(1),  015011 (2018)

\bibitem{zlocha2019improving}
Zlocha, M., Dou, Q., Glocker, B.: Improving retinanet for ct lesion detection
  with dense masks from weak recist labels. arXiv preprint arXiv:1906.02283
  (2019)

\end{thebibliography}
\end{document}